\pdfoutput=1

\documentclass[11pt]{article}

\usepackage[preprint]{acl}

\usepackage{times}
\usepackage{latexsym}
\usepackage{algpseudocode}
\usepackage{algorithm}

\usepackage[T1]{fontenc}

\usepackage[utf8]{inputenc}

\usepackage{microtype}

\usepackage{inconsolata}

\usepackage{graphicx}
\usepackage[acronym]{glossaries}

\usepackage{inconsolata}

\usepackage{multicol}
\usepackage{booktabs}
\usepackage{makecell}
\usepackage{amsmath, amssymb}
\usepackage{multirow}
\usepackage{mathtools}
\usepackage{xcolor}
\usepackage{enumitem}
\usepackage{float}
\usepackage{graphicx}
\usepackage{upgreek}
\usepackage{seqsplit}
\usepackage{color,soul}
\usepackage{arydshln}
\usepackage{placeins}
\usepackage{pifont}
\usepackage{amssymb}
\usepackage{bbm}

\usepackage{amsmath,amsfonts,bm}

\def\eqref#1{equation~\ref{#1}}

\def\1{\bm{1}}

\def\rva{{\mathbf{a}}}

\def\rvc{{\mathbf{c}}}
\def\rvd{{\mathbf{d}}}

\def\rvn{{\mathbf{n}}}

\def\rvq{{\mathbf{q}}}
\def\rvr{{\mathbf{r}}}
\def\rvs{{\mathbf{s}}}

\DeclareMathAlphabet{\mathsfit}{\encodingdefault}{\sfdefault}{m}{sl}
\SetMathAlphabet{\mathsfit}{bold}{\encodingdefault}{\sfdefault}{bx}{n}

\usepackage{caption}
\usepackage{subcaption}
\title{Agentic Medical Knowledge Graphs Enhance Medical Question Answering: Bridging the Gap Between LLMs and Evolving Medical Knowledge}

\author{Mohammad R. Rezaei$^{1,2\boldsymbol{*}}$, Reza Saadati Fard$^3$,Jayson L. Parker$^1$, Rahul G. Krishnan$^{1,2}$, Milad Lankarany$^{1}$\\ 
        $^1$ University of Toronto \\
        $^2$ Vector Institute\\
        $^3$ Worcester Polytechnic Institute\\
        \textsuperscript{*}\texttt{mr.rezaei@mail.utoronto.ca}}

\usepackage{pifont}
\newcommand{\yesmarker}{\textcolor{black}{\ding{51}}} 
\newcommand{\nomarker}{\textcolor{red}{\ding{55}}}   
\begin{document}
\newacronym{myrag}{AMG-RAG}{Agentic Medical Graph-RAG}
\newacronym{kg}{KG}{Knowledge Graph}
\newacronym{mkg}{MKG}{Medical Knowledge Graph}

\newacronym{llm}{LLM}{Large Language Model}
\newacronym{cot}{CoT}{Chain-of-Thought}
\newacronym{qa}{QA}{Question Answering}
\newacronym{rag}{RAG}{Retrieval Augmented Generation }
\maketitle
\begin{abstract}
\glspl{llm} have greatly advanced medical \gls{qa} by leveraging vast clinical data and medical literature. However, the rapid evolution of medical knowledge and the labor-intensive process of manually updating domain-specific resources can undermine the reliability of these systems. We address this challenge with \gls{myrag}, a comprehensive framework that automates the construction and continuous updating of \glspl{mkg}, integrates reasoning, and retrieves current external evidence from the \glspl{mkg} for medical \gls{qa}.
Evaluations on the MEDQA and MEDMCQA benchmarks demonstrate the effectiveness of \gls{myrag}, achieving an F1 score of 74.1\% on MEDQA and an accuracy of 66.34\% on MEDMCQA—surpassing both comparable models and those 10 to 100 times larger. By dynamically linking new findings and complex medical concepts, \gls{myrag} not only boosts accuracy but also enhances interpretability for medical queries, which has a critical impact on delivering up-to-date, trustworthy medical insights (GitHub: \url{https://github.com/MrRezaeiUofT/AMG-RAG}).
\end{abstract}

\glsresetall
\section{Introduction}

Medical knowledge is growing at an unprecedented rate: every day brings new research findings, revised clinical guidelines, and updated treatment protocols.  Recent work shows that \glspl{llm} can already harness this ever‑expanding corpus for medical \gls{qa} \cite{nazi2024large,liu2023utility}.  
\begin{figure}[t]
  \centering
  \includegraphics[width=.9\columnwidth]{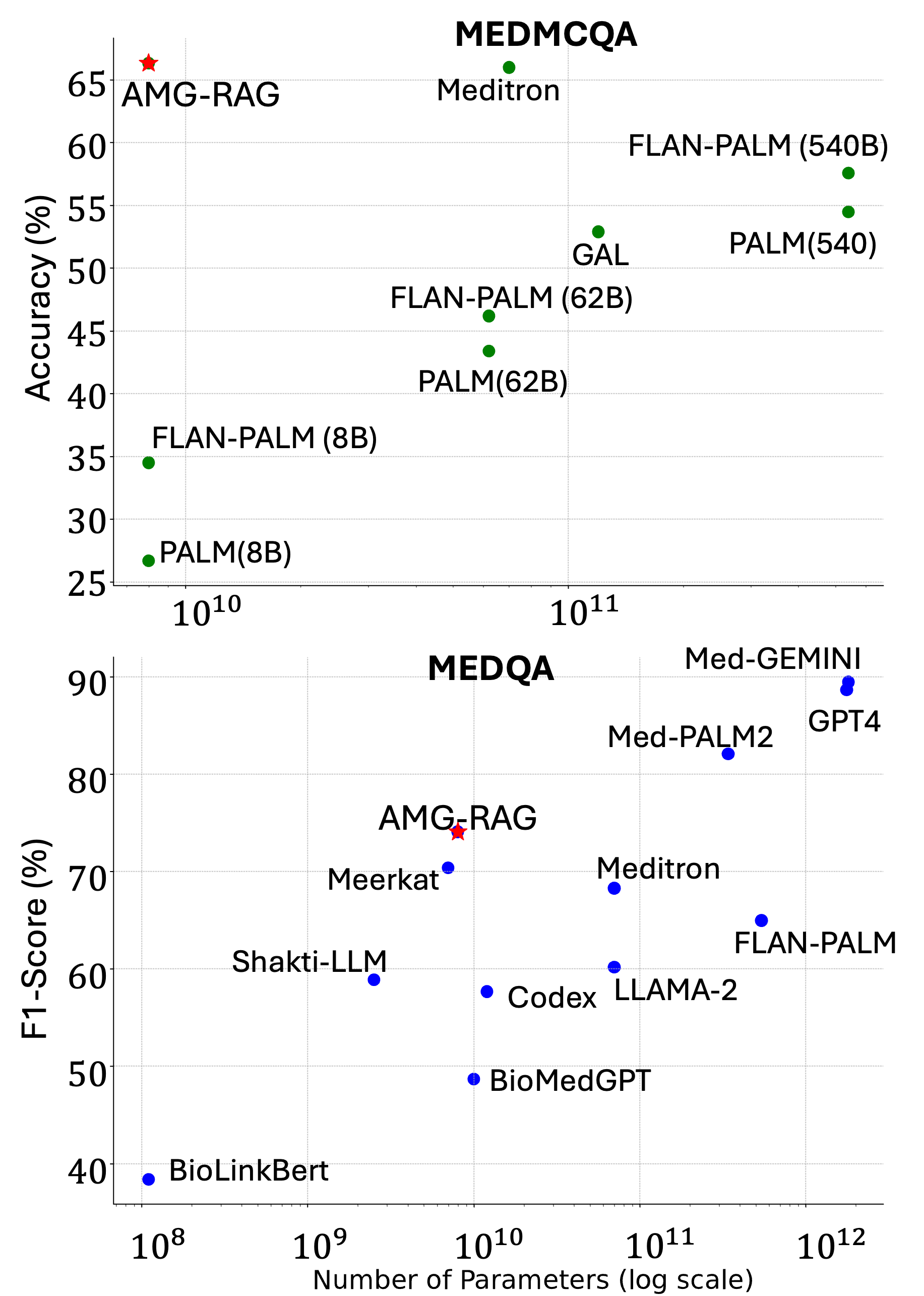}
  \caption{Performance versus parameter count on the MEDQA and MEDMCQA benchmarks.  Our system, \gls{myrag}, attains an F1 of 74.1\,\% on MEDQA and an accuracy of 66.34\,\% on MEDMCQA, outperforming models that contain 10–100× more parameters.  See Tables~\ref{tab:medqa_comparison} and \ref{tab:medmcqa_comparison} for details.}
  \label{fig:performance_vs_size}
\end{figure}

Despite their promise, \glspl{llm} face two persistent challenges.  First, they must remain \emph{factually current} in a field where knowledge can become obsolete almost overnight \cite{rohanian2024exploring,yu2024large}.  Second, they must correctly model the intricate relationships among medical entities.  \gls{kg} provide a structured and interconnected view that supports nuanced reasoning \cite{huang2021knowledge}, yet creating and maintaining them by hand is costly—especially in medicine, where new insights rapidly invalidate older facts \cite{yang2024kg}.

We introduce an automated framework that constructs and continuously refines \glspl{mkg} for \gls{qa}.  Our \gls{llm}‑driven agents, assisted by domain‑specific search tools, generate graph entities enriched with metadata, confidence scores, and relevance indicators.  This automation sharply reduces manual curation while keeping the graph aligned with the latest discoveries.  In contrast to \gls{rag} systems that rely solely on vector similarity \cite{lewis2020retrieval}, our graph‑centric retrieval leverages explicit relationships to synthesise information across domains such as drug interactions, clinical trials, patient histories, and guidelines.

The \gls {myrag} combines dynamically synthesised \gls{mkg} with multi‑step reasoning, guided by confidence scores and adaptive traversal strategies \cite{trivedi2022interleaving}.  This design yields more accurate and complete answers without incurring additional fine‑tuning or inference costs.
On the MEDQA and MEDMCQA benchmarks—both of which test evidence retrieval, complex reasoning, and multiple‑choice comprehension—\gls{myrag} achieves an F1 of 74.1\% and an accuracy of 66.34\%, respectively (Fig.~\ref{fig:performance_vs_size}).  These results surpass those of similarly sized \gls{rag} approaches and even much larger state‑of‑the‑art models, underscoring the benefits of a dynamically evolving \gls{mkg} for medical \gls{qa}.  Our findings highlight the potential of automated, relationally enriched knowledge retrieval to enhance clinical decision‑making by delivering timely and trustworthy insights \cite{zhou2023survey}.

\paragraph{Contributions.}Our contributions are threefold:
\begin{enumerate}
   \item We developed a autonomous search and graph-building process powered by specialized \gls{llm} agents that continuously generate and refine \glspl{mkg} through integrated workflows using search engines and medical textbooks.
   \item Our system embeds confidence scoring mechanisms that explicitly model information uncertainty, providing transparent reliability assessments for medical information.
   \item We created an adaptive graph traversal system that transcends traditional retrieval methods, enabling dynamic contextualization of medical knowledge.
\end{enumerate}

\section{Related Work}
Medical \gls{qa} has progressed through three complementary lines of research:  
{(i) domain‑specific language models,} {(ii) retrieval‑augmented generation,} and {(iii) knowledge‑graph reasoning}. 
\paragraph{Domain‑specific language models.}
BioBERT \cite{lee2020biobert}, PubMedBERT \cite{gu2021domain}, and MedPaLM \cite{singhal2023large} adapt transformer pre‑training to biomedical corpora, delivering strong gains on entity recognition, relation extraction, and multiple‑choice \gls{qa} \cite{nazi2024large,liu2023utility}.  Yet, even these specialised models struggle to synthesise multi‑hop relations (e.g.\ rare comorbidities or drug–gene interactions) and must be re‑trained to absorb new discoveries \cite{rohanian2024exploring,yu2024large}.
\paragraph{Retrieval‑Augmented Generation (RAG).}
RAG pipelines couple an \gls{llm} with an external evidence retriever, injecting fresh context at inference time \cite{lewis2020retrieval}.  Vendi-\gls{rag} \cite{rezaei2025vendirag} and MMED‑\gls{rag}\cite{xia2024mmed} extend this idea to biomedical and multimodal sources, respectively.  Chain‑of‑Thought (CoT) prompting further boosts reasoning: IRCoT \cite{trivedi2022interleaving} interleaves iterative retrieval with step‑wise justification.  Gemini’s long‑context model recently pushed MedQA scores beyond GPT‑4 \cite{saab2024capabilities}.  Nevertheless, most \gls{rag} systems rely on static vector stores and cannot \emph{explain} answers in terms of explicit biomedical relations.

\paragraph{Knowledge‑graph reasoning.}
KG‑Rank \cite{huang2021knowledge} and related work such as KG-\gls{rag}\cite{sanmartin2024kg} harness ontologies to re‑rank evidence or enforce logical constraints, improving factual consistency in long‑form \gls{qa} \cite{yang2024kg}.  However, constructing and curating a high‑coverage, up‑to‑date \gls{mkg} remains labour‑intensive, limiting scalability and freshness.

\glsresetall
\section{Method}
We propose our framework, \gls{myrag}, bridges these threads by \emph{dynamically} generating a confidence‑scored \gls{mkg} that is \emph{tightly coupled} to a \gls{rag}+CoT pipeline. \gls{myrag} features autonomous \gls{kg} evolution through \gls{llm} agents extracting entities and relations from live sources with provenance tracking; graph‑conditioned retrieval that maps queries onto the \gls{mkg} to guide evidence selection; and reasoning over structured context where the answer generator utilizes both textual passages and traversed sub‑graphs for transparent, multi‑hop reasoning.

\subsection{Retrieval Augmented Generation (RAG)}
\gls{rag} is a framework designed to enhance \gls{qa} by integrating relevant external knowledge into the generation process.In the \gls{rag} approach, the retriever fetches a fixed number of relevant documents, $\{\rvd_1, \rvd_2, \ldots, \rvd_n\} \in \mathbf{D}$, based on the query $\rvq$. Here, $\mathbf{D}$ represents the set of all domain-specific documents utilized. These documents are concatenated and passed directly to a \gls{llm}-based text generator, $G$, which produces the answer $\hat{\rva}$:
\[
\hat{\rva} = G(\rvq, \{\rvd_1, \ldots, \rvd_n\}).
\]
This approach is simple and computationally efficient but may struggle with domain-specific or complex queries that require additional supporting evidence.

\paragraph{RAG with Chain-of-Thought (CoT).}
Enhancing \gls{rag}'s performance can be achieved by integrating intermediate reasoning steps prior to producing the final response. The generator produces a chain of thought, $\rvc$, which serves as an explicit reasoning trace:
\[
\{\rvd_1, \ldots, \rvd_k\} = \text{Retriever}(\rvq; \mathbf{D}),\] \[\rvc = G(\rvq, \{\rvd_1, \ldots, \rvd_k\}),\quad \hat{\rva} = G(\rvc).
\]
This step-by-step approach enhances reasoning and interpretability, leading to improved accuracy in multi-hop reasoning tasks.

\paragraph{RAG with Search.}
The \glspl{rag}'s performance can improved further by incorporating additional related documents retrieved from external sources, such as the internet, through a search tool. This variant integrates external search capabilities into the retrieval process. For a query \(\rvq\), the retriever's results are combined with those from external search engines, providing more comprehensive evidence for the \gls{llm} to generate a response:  
\[
\{\rvd'_1, \ldots, \rvd'_m\} = \text{Search}(\rvq; \mathbf{D'}),
\]  
\[
\hat{\rva} = G(\rvq, \{\rvd_1, \ldots, \rvd_n, \rvd'_1, \ldots, \rvd'_m\}).
\]  
This additional search step significantly enhances performance, particularly in scenarios that require access to extensive and diverse knowledge.
\begin{figure*}[t]
    \centering
\includegraphics[width=.93\textwidth]{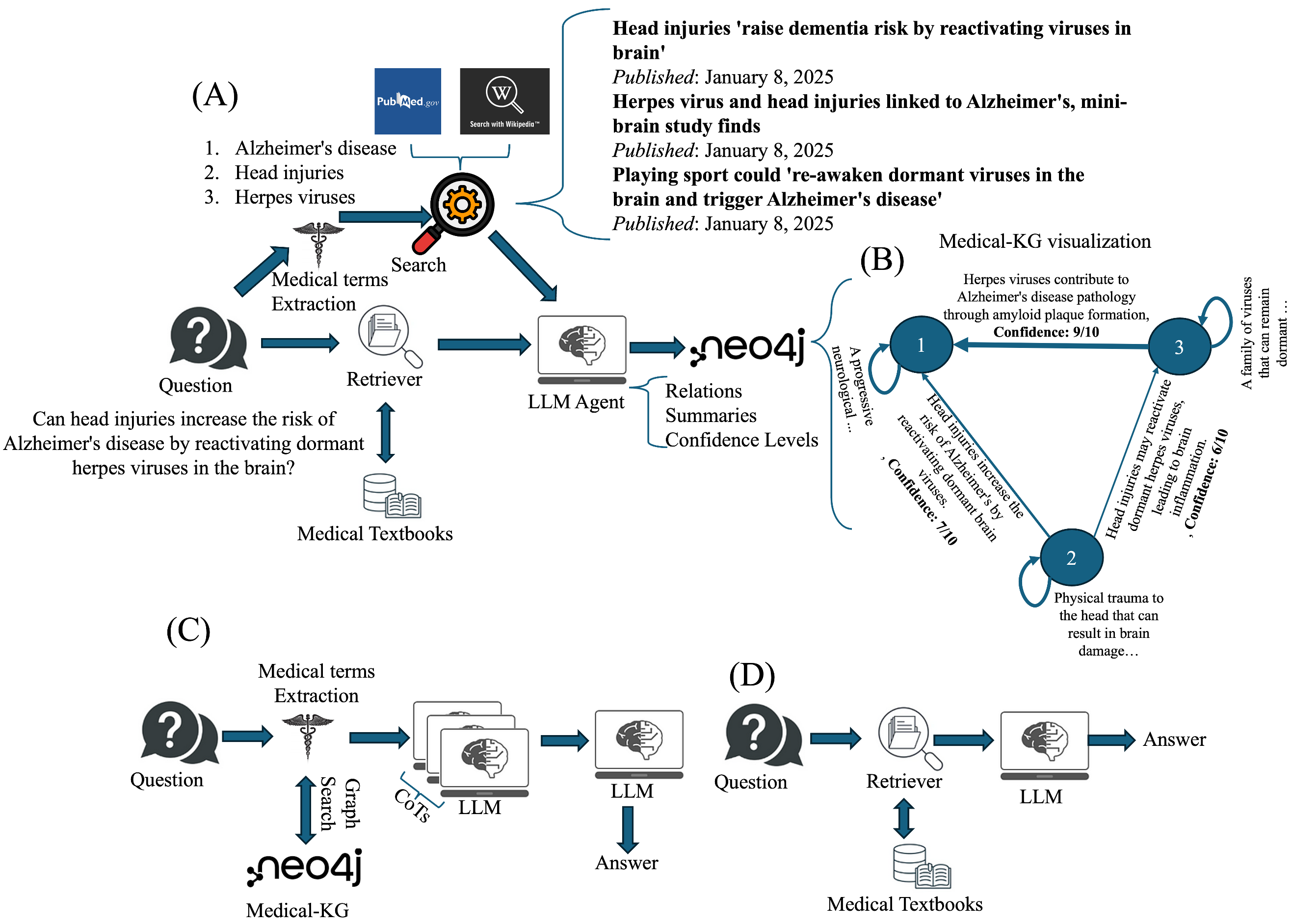}
    \caption{Model Schema. A) The pipeline for creating the \gls{mkg} using search tools and an \gls{llm} agent. B) An example of the generated \gls{mkg} in Neo4J, illustrating nodes and relationships derived from search results and contextual information. Our model successfully retrieved and utilized recent knowledge to accurately answer a medical question, highlighting the practical benefit of continuously updating the knowledge graph. Furthermore, we extended this evaluation by providing additional examples retrieved by our system using recent publications in Table~\ref {tab:amg_rag_examples}. C) The \gls{myrag} pipeline. D) A simplified \gls{rag} pipeline. }
    \label{fig:model_schema}
\end{figure*}
\subsection{Medical QA with AMG-RAG}
\noindent
\begin{algorithm}
\small
\caption{KG-Based QA Inference Pipeline}
\begin{algorithmic}[1]
\Require Query $q$, Knowledge Graph $KG$, Confidence Threshold $\tau$, Max Iterations $N$
\Ensure Final Answer $\hat{\rva}$ with Confidence $s$
\State Extract medical terms: $\{\rvn_1, \rvn_2, \ldots, \rvn_m\} \gets \text{ExtractTerms}(\rvq)$
\State Initialize reasoning traces: $C \gets \emptyset$
\State Initialize confidence: $\rvs_i \gets 1.0$ for all terms $\rvn_i$
\For{$i = 1$ to $m$} \Comment{Iterate over extracted terms}
    \State \textbf{Explore KG:} Retrieve relevant nodes $\{\rvd_j\}$ and relationships $\rvr_{ij}$ for $\rvn_i$
    \For{each child node $\rvn_j$ of $\rvn_i$ in $KG$}
        \State Compute child confidence: $s_j \gets s_i \cdot \rvr_{ij}$
        \If{$s_j \geq \tau$}
            \State Include $\rvn_j$ in exploration set
        \EndIf
    \EndFor
    \State \textbf{Generate Reasoning Trace:} $\rvc_i \gets \text{LLM}(\rvn_i, \{\rvd_j\})$
    \State Add $\rvc_i$ to reasoning traces: $C \gets C \cup \{\rvc_i\}$
\EndFor
\State \textbf{Synthesize Answer:} $\hat{a}, \hat{s} \gets G(C)$
\State \Return $\hat{\rva}, \hat{s}$ \Comment{Return final answer with confidence}
\end{algorithmic}
\label{alg:kg-qa-pipeline}
\end{algorithm}
In scenarios requiring domain expertise, such as medical or scientific \gls{qa}, traditional methods often fail due to their inability to capture intricate domain-specific relationships or handle ambiguous queries. \gls{kg}-driven approaches overcome these challenges by integrating explicit relationships and structured knowledge representations. This marks a significant advancement in intelligent \gls{qa} systems, ensuring robustness and scalability across various applications.

The suggested \gls{myrag} framework dynamically creates a \gls{mkg} and incorporates sophisticated reasoning abilities, overcoming the shortcomings of traditional methods. Our system utilizes structured medical knowledge and reasoning, ensuring flexibility to accommodate new data.

The \gls{myrag} pipeline begins with question parsing, where an \gls{llm} agent extracts medical terms $\{\rvn_1, \rvn_2, \ldots, \rvn_m\}$ from the user query $\rvq$:
\[
\{\rvn_1, \rvn_2, \ldots, \rvn_m\} = \text{LLM}(\rvq, M), \quad m \leq M.
\]

During node exploration, the system queries the \gls{kg} for each term $\rvn_i$, applying a confidence threshold that filters relationships based on their reliability scores. The system propagates confidence through the \gls{kg} by computing child confidence as:
\[
s(\rvn_j) = s(\rvn_i) \cdot s(\rvr_{ij}), \quad \forall j \in \text{children of } i.
\]

Our framework supports both breadth-first and depth-first exploration strategies, enabling flexible knowledge traversal based on query characteristics. The exploration continues until either cumulative confidence meets threshold $\tau$ or document limit $M$ is reached, ensuring comprehensive yet focused information gathering.

The chain-of-thought generation phase synthesizes reasoning traces $\rvc_i$ for each entity by integrating information from connected nodes:
\[
\rvc_i = \text{LLM}(\rvn_i, \{\rvd(\rvn_j) \mid j \in \text{connected nodes}\}).
\]

Finally, answer synthesis aggregates these reasoning traces to produce the final output $\hat{\rva}$ with an associated confidence score:
\[
\hat{\rva}, \hat{s} = G(\{\rvc_1, \rvc_2, \ldots, \rvc_m\}).
\]
This approach ensures that answers are comprehensive, interpretable, and anchored in reliable medical knowledge.
\subsection{Dynamic Generation of the Medical Knowledge Graph}
The development of the \gls{mkg} marks a pivotal advancement within our \gls{myrag} framework, facilitating organized reasoning through the use of dynamically synthesized knowledge representations. Unlike traditional static knowledge bases, our method allows the graph structure to adapt continuously in response to new queries and the evidence they uncover. Furthermore, it assigns a confidence score to each edge in the graph, indicating the reliability of each relationship, which is crucial given the uncertainty inherent in medical evidence.

\paragraph{Node Extraction.}
Our Medical Entity Recognizer (MER) agent identifies domain-specific terms within user queries, establishing them as foundational nodes $\{\rvn_1, \rvn_2, \ldots, \rvn_m\}$ in the knowledge graph. For each identified entity, specialized medical search tools retrieve contextual descriptions $\rvd(\rvn_i)$:
\[
\rvd(\rvn_i) = \text{Search}(\rvn_i; \text{knowledge source}).
\]
These descriptions provide rich semantic context for the knowledge graph, ensuring accurate representation of medical concepts and their attributes.

\paragraph{Relationship Inference.}
The power of our approach lies in its ability to dynamically infer relationships between medical entities. An \gls{llm} agent analyzes pairs of nodes $(\rvn_i, \rvn_j)$ to determine potential relationships $\rvr_{ij}$ and quantify their reliability:
\[
\rvr_{ij}, \rvs_{ij} = \text{LLM}(\rvd(\rvn_i), \rvd(\rvn_j)).
\]
This process generates not just connection types (e.g., causation, correlation, contraindication) but also confidence scores that propagate through subsequent reasoning steps, enabling evidence-weighted inference.

\paragraph{Knowledge Graph Construction.}
The resulting \gls{mkg} integrates extracted entities, their descriptions, inferred relationships, and confidence metrics into a cohesive structure. This graph serves as both a repository of medical knowledge and a computational framework for reasoning. By attaching reliability scores to every edge, downstream components can appropriately weight evidence during inference, enhancing both accuracy and explainability. The graph's dynamic nature allows it to continuously incorporate new information and refine existing connections, addressing the staleness issues that plague traditional knowledge-based systems.
\section{Experiments}
The \textbf{MEDQA} dataset is a free-form, multiple-choice open-domain \gls{qa} dataset specifically designed for medical \gls{qa}. Derived from professional medical board exams, this dataset presents a significant challenge as it requires both the retrieval of relevant evidence and sophisticated reasoning to answer questions accurately. Each question is accompanied by multiple-choice answers that demand a deep understanding of medical concepts and logical inference, often relying on evidence found in medical textbooks. For this study, the test partition of the MEDQA dataset, comprising approximately 1,200 samples, was used \cite{jin2021disease}.

The \textbf{MedMCQA} dataset is another multiple-choice question-answering dataset tailored for medical \gls{qa}. Unlike MEDQA, which is derived from board exam questions, MedMCQA offers a broader variety of question types, encompassing both foundational and clinical knowledge across diverse medical specialties. In this study, the MedMCQA development set, containing approximately 4,000 questions, was used to benchmark against other models \cite{pmlr-v174-pal22a}.

This study employed the MEDQA and MedMCQA datasets to benchmark and evaluate medical \gls{qa} systems. These datasets serve as challenging testbeds for open-domain \gls{qa} tasks due to their demands for multi-hop reasoning and the integration of domain-specific knowledge. The relevance of MEDQA in the real world, together with the diverse question styles and extensive development set of MedMCQA make them ideal for advancing the development of robust \gls{qa} models capable of addressing medical inquiries. We utilize \textit{GPT-4o-mini} as the backbone of the implementation for both \gls{mkg} and \gls{myrag}, leveraging its capabilities with approximately \(\sim 8B\) parameters. This model serves as the core component, enabling advanced reasoning, \gls{rag}, and structured knowledge integration.

\subsection{Medical Knowledge Graph}

To address the challenges of inaccurate knowledge updating—such as those stemming from noisy retrieval results or \gls{llm} hallucinations—our \gls{myrag} introduces a robust and dynamic approach to \gls{mkg} construction. This is particularly critical in healthcare applications, where the absence of error detection and correction mechanisms in automated \gls{kg} generation can compromise system reliability.

The dynamic update mechanism encompasses strategies resilient to errors, defined by the confidence level of the medical information retained within the \gls{mkg} and among nodes $i$ and $j$, denoted as $\rvs_{ij}$. This approach facilitates monitoring and reduces the spread of erroneous information during the refinement or reasoning stages. These protective measures allow the system to identify and rectify inconsistencies that may arise from external retrieved information.

The \gls{mkg} is dynamically constructed for each question by integrating search items, contextual information, and relationships extracted from medical textbooks and search tools, including Wikipedia (Wiki-MKG) and PubMed (PubMed-MKG) queries. The ablation in Table \ref{tab:medqa_results} demonstrates that the created \gls{mkg} based on PubMed (PubMed-MKG) is more effective in enhancing the performance of the \gls{myrag}. This data is processed and structured within a Neo4j database. Key innovations in the knowledge graph include:

\begin{enumerate}
    \item \textbf{Dynamic Node and Relationship Creation}: Nodes are instantiated based on retrieved entities and search terms, while relationships are constructed using predefined semantic templates aligned with medical ontologies.

    \item \textbf{Bidirectional Relationships}: The graph includes both forward and reverse relationships between nodes to allow flexible traversal and comprehensive context understanding.

    \item \textbf{Confidence-Based Relevance Scoring}: Each relationship is enriched with textual annotations and a quantitative confidence score that measures the reliability of the connection. This confidence score enables the system to down-rank or filter out uncertain associations, thereby mitigating the effects of noisy retrievals.

    \item \textbf{Summarization with Reliability Indicators}: Each search item is paired with a concise summary derived from contextual sources. These summaries are accompanied by confidence scores that indicate their trustworthiness, allowing nuanced uncertainty modeling.

    \item \textbf{Thresholding for Quality Control}: In our experiments, we applied a confidence threshold of 8 (on a 10-point scale) to retain only high-reliability nodes and edges. This value was empirically found to yield the best results in benchmark performance.

    \item \textbf{Integration with Neo4j}: The complete graph is stored in a Neo4j database, leveraging its powerful graph query engine for efficient retrieval and analysis during inference.
\end{enumerate}

A partial visualization of the \gls{mkg} structure is shown in Figure~\ref{fig:model_schema}.B. Additional complete examples with retrieved papers are provided in Table~\ref{tab:amg_rag_examples}. This \gls{mkg} forms the core knowledge source for the \gls{myrag} inference pipeline.

As discussed in Appendix~\ref{app:mkd-analysis}, extensive validation was conducted through both human and machine evaluators. Clinical experts verified the correctness of the knowledge graph, and expert \glspl{llm} such as GPT-4 achieved high accuracy (e.g., 9/10) in validating the extracted knowledge.  These results underscore the \gls{mkg}'s ability to support reliable and explainable medical reasoning within \gls{myrag}.

The knowledge graph creation process in \gls{myrag} operates independently from the \gls{qa} process, allowing for continuous background updates of the \gls{mkg} via search tools such as PubMedSearch or WikiSearch. This approach significantly reduces latency during question answering since the system frequently retrieves information from the pre-populated \gls{mkg} rather than performing new searches. By maintaining an updated \gls{mkg}, \gls{myrag} achieves a balanced minimum dependency on computational resources and search tools during the test phase.

Despite having only 8B parameters, it delivers competitive results compared to much larger models like Med-Gemini (1800B) and GPT-4 (1760B). Even in worst-case scenarios where relevant information is absent from the \gls{mkg}, the additional search cost is still significantly lower than the resource requirements of much larger models.
\subsection{Performance Comparison}
\begin{table*}[h!]
\small
\centering

\renewcommand{\arraystretch}{0.9}
\resizebox{\textwidth}{!}{%
\begin{tabular}{@{}lcccccc@{}}
\toprule
\textbf{Model}            & \textbf{Model Size} & \textbf{Acc. (\%)} & \textbf{F1 (\%)} & \textbf{Fine-Tuned} & \textbf{Uses CoT} & \textbf{Uses Search} \\
\midrule
Med-Gemini \cite{saab2024capabilities}               & $\sim$1800B                                     & 91.1                   & 89.5                   & \yesmarker          & \yesmarker        & \yesmarker          \\
GPT-4 \cite{nori2023can}                     & $\sim$1760B                                     & 90.2                   & 88.7                   & \yesmarker          & \yesmarker        & \yesmarker          \\
Med-PaLM 2 \cite{singhal2025toward}               & $\sim$340B                                     & 85.4                   & 82.1                   & \yesmarker          & \yesmarker        & \nomarker           \\
Med-PaLM 2 (5-shot)       & $\sim$340B                                     & 79.7                   & 75.3                   & \nomarker           & \yesmarker        & \nomarker           \\
\gls{myrag}                 & $\sim$8B                                     & 73.9                   & 74.1                   & \nomarker          & \yesmarker        & \yesmarker           \\
Meerkat\cite{kim2024small}              & 7B                                       & 74.3                   & 70.4                   & \yesmarker          & \yesmarker        & \nomarker           \\
Meditron \cite{chen2023meditron}                 & 70B                                      & 70.2                   & 68.3                   & \yesmarker          & \yesmarker        & \yesmarker          \\
Flan-PaLM  \cite{singhal2023large}               & 540B                                     & 67.6                   & 65.0                   & \yesmarker          & \yesmarker        & \nomarker           \\
LLAMA-2 \cite{chen2023meditron}                  & 70B                                      & 61.5                   & 60.2                   & \yesmarker          & \yesmarker        & \nomarker           \\
Shakti-LLM \cite{shakhadri2024shakti}               & 2.5B                                     & 60.3                   & 58.9                   & \yesmarker          & \nomarker         & \nomarker           \\
Codex 5-shot CoT  \cite{lievin2024can}        & --                                     & 60.2                   & 57.7                   & \nomarker           & \yesmarker        & \yesmarker          \\
BioMedGPT  \cite{luo2023biomedgpt}               & 10B                                      & 50.4                   & 48.7                   & \yesmarker          & \nomarker         & \nomarker           \\
BioLinkBERT (base)  \cite{singhal2023large}      & --                                     & 40.0                   & 38.4                   & \yesmarker          & \nomarker         & \nomarker           \\
\bottomrule
\end{tabular}%
}
\caption{Comparison of LLM models on the MEDQA Benchmark. Additional comparison with RAGs are provided in Table \ref{app:mkd-analysis}}
\label{tab:medqa_comparison}
\end{table*}

Table~\ref{tab:medqa_comparison} presents a comprehensive comparison of state-of-the-art language models on the MEDQA benchmark. The results highlight the critical role of advanced reasoning strategies in achieving higher performance, such as CoT reasoning and the integration of search tools. While larger models like Med-Gemini and GPT-4 achieve the highest accuracy and F1 scores, their performance comes at the cost of significantly larger parameter sizes. These models exemplify the power of scaling combined with sophisticated reasoning and retrieval techniques.

Significantly, \gls{myrag}, despite having just 8 billion parameters, attains an F1 score of 74.1\% on the MEDQA benchmark, surpassing models like Meditron, which possess 70 billion parameters without needing any fine tuning. This highlights \gls{myrag}'s exceptional efficiency and proficiency in utilizing CoT reasoning and external evidence retrieval. The model leverages tools such as PubMedSearch and WikiSearch to dynamically integrate domain-specific knowledge dynamically, thereby improving its ability to address medical questions. Examples of \gls{qa} interactions, including detailed search items and reasoning for question samples, are provided in Appendix \ref{app:examples-medqa}. These examples are organized in Tables \ref{table:search_guidance_1}, \ref{table:search_guidance_2}, \ref{table:search_guidance_3}, and \ref{table:search_guidance_4}, drawn from the MEDQA benchmark.

On the MedMCQA benchmark, as shown in Table~\ref{tab:medmcqa_comparison}, \gls{myrag} achieves an accuracy of 66.34\%, even outperforming larger models like Meditron-70B and better than Codex 5-shot CoT. This result underscores \gls{myrag}'s adaptability and robustness, demonstrating that it can deliver competitive performance even against significantly larger models. Its ability to maintain high accuracy on diverse datasets further highlights the effectiveness of its design, which combines CoT reasoning with structured knowledge graph integration and retrieval mechanisms.

\begin{table}[h!]
\small
\centering

\renewcommand{\arraystretch}{0.9}

\begin{tabular}{@{}lcc@{}}
\toprule
\textbf{Model}                              & \textbf{Model Size}           & \textbf{ Acc. (\%)} \\
\midrule
\gls{myrag}                 & $\sim$8B                                     & \textbf{66.34}\\
Meditron \citep{chen2023meditron}                     & 70B                          & 66.0 \\

Codex 5-shot \citep{lievin2024can}                            & --                           & 59.7 \\
VOD \citep{lievin2023variational}                           & --                           & 58.3 \\
Flan-PaLM \citep{singhal2022large}                        & 540B                         & 57.6 \\
PaLM                       & 540B                         & 54.5 \\

GAL                        & 120B                         & 52.9 \\
PubmedBERT \citep{gu2021domain}                & --                           & 40.0 \\
SciBERT \citep{pal2022medmcqa}              & --                           & 39.0 \\
BioBERT \citep{lee2020biobert}                  & --                           & 38.0 \\
BERT \citep{devlin2018bert}                  & --                           & 35.0 \\
\bottomrule
\end{tabular}%
\caption{Comparison of Models on the MedMCQA.}
\label{tab:medmcqa_comparison}
\end{table}

Overall, \gls{myrag}'s results on MEDQA and MedMCQA benchmarks solidify its position as a highly efficient and effective model for medical \gls{qa}. By leveraging reasoning, dynamically generated \gls{mkg}, and external knowledge sources, \gls{myrag} not only closes the gap with much larger models but also sets a new standard for performance among smaller-sized models.
\paragraph{Impact of Search Tools on MKG creation and CoT Reasoning on AMG-RAG Performance.}

\begin{figure}[t]
    \centering
    \includegraphics[width=\columnwidth]{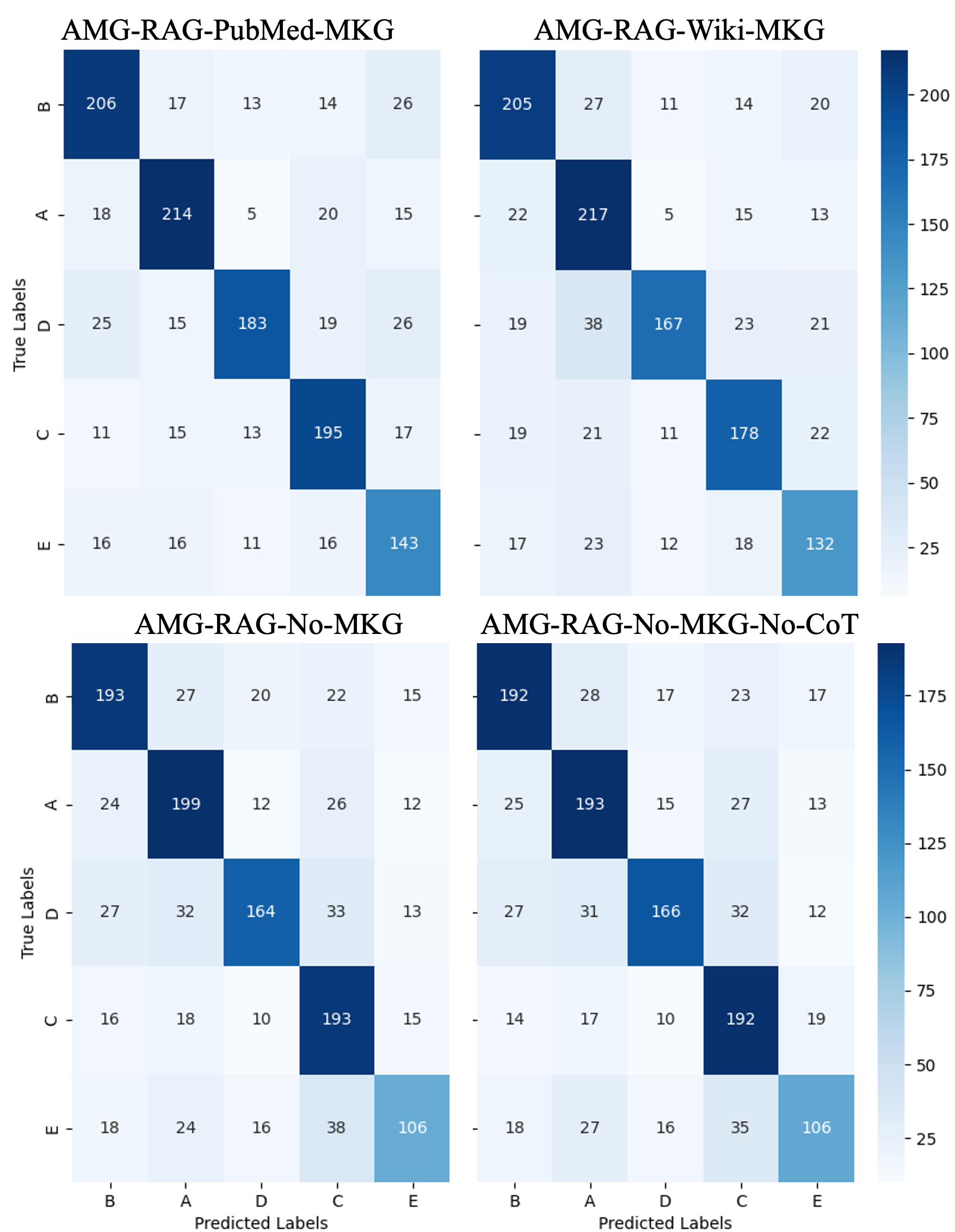}
    \caption{Confusion matrix for \gls{myrag} with and without CoT and Knowledge Graph integration on the MEDQA dataset.}
    \label{fig:cot_kg_comparison}
\end{figure}

Figure~\ref{fig:cot_kg_comparison} and Table~\ref{tab:medqa_results} demonstrate the effect of integrating different search tools for creating the \gls{mkg} on the performance of the \gls{myrag} system applied to the MEDQA benchmark. Incorporating these external retrieval capabilities significantly enhances both accuracy and F1 scores, as they allow the model to access relevant and up-to-date evidence critical for answering complex medical questions. Among the two search tools for creating the \gls{mkg}, PubMed-MKG consistently outperforms Wiki-MKG, likely due to its focused, domain-specific content that aligns closely with the specialized nature of medical \gls{qa} tasks.

In addition to the integration of the dynamical \gls{mkg}, the reasoning module plays a pivotal role in performance. As highlighted in Figure~\ref{fig:cot_kg_comparison}, ablating either CoT or \gls{mkg} integration causes a considerable degradation in accuracy and F1 score. This demonstrates that structured multi-hop reasoning and medical knowledge grounding through the \gls{mkg} are indispensable for the system's ability to deliver accurate and evidence-based answers."

\paragraph{Comparison Against Traditional RAG Models.}

Table~\ref{tab:medqa_results} presents a comprehensive comparison of various RAG models evaluated on the MEDQA benchmark. This includes models with different retrieval mechanisms and model sizes, enabling a head-to-head evaluation of \gls{myrag} with other state-of-the-art baselines such as Self-RAG \citep{asai2023self}, HyDE \citep{gao2022precise}, GraphRAG \citep{edge2024local}, and MedRAG \citep{zhao2025medrag}. The results clearly show that \gls{myrag} configured with the PubMed-\gls{mkg} and an 8B LLM backbone achieves the highest accuracy of 73.92\%, surpassing all competing models. Notably, ablation results indicate that removing search functionality or CoT reasoning significantly degrades accuracy (dropping to 67.16\% and 66.69\%, respectively), confirming the essential role of structured retrieval and reasoning components in complex question answering. Other baseline models such as Gemini-pro and PMC-LLaMA demonstrate weaker performance, further validating the efficacy of domain-aware retrieval and reasoning modules proposed in \gls{myrag}. Importantly, the domain specificity and freshness of PubMedSearch provide a significant advantage in retrieving relevant knowledge that general-purpose search modules often fail to deliver.

\begin{table}[h!]
\centering
\small
\begin{tabular}{lcc}
\hline
\textbf{Model} & \textbf{Size} & \textbf{Accuracy (\%)} \\
\hline
\gls{myrag} & PubMed-MKG-8B & \textbf{73.92} \\
            & Wiki-MKG-8B & 70.62 \\
            & No-MK-8B & 67.16 \\
            & No-MKG \& CoT-8B & 66.69 \\
\hline
Self-RAG & 8B & 67.32 \\
         & HyDE-8B & 68.32 \\
\hline
RAG & Gemini-pro & 64.5 \\
             & 70B & 56.2 \\
             & 8B  & 64.3 \\
\hline
GraphRAG & Gemini-pro & 65.1 \\
                  & 70B & 55.1 \\
                  & 8B  & 64.8 \\
\hline
MedRag & 70B & 49.57 \\
       & 13B & 42.58 \\
\hline
PMC-LLaMA & 13B & 44.38 \\
\hline
\end{tabular}
\caption{Comparison of MEDQA accuracy across various RAG models and retrieval strategies.}
\label{tab:medqa_results}
\end{table}

\paragraph{Comparison Against LLM Backbones.}

In addition to evaluating different retrieval strategies, we assess how the choice of LLM backbone influences performance in Table~\ref{tab:medqa_llm_results}. This comparison highlights that \gls{myrag} built on GPT4o-mini with PubMed-MKG achieves the best performance (73.92\%). In contrast, performance declines when switching to LLaMA 3.1 or Mixtral, even when using the same retrieval pipeline. These results reinforce the importance of synergy between the language model and the retrieval mechanism. Larger models do not necessarily guarantee higher accuracy—domain alignment and reasoning ability, such as that of GPT4o-mini, are crucial for success on high-stakes tasks like medical QA.

\begin{table}[h!]
\small
\centering
\begin{tabular}{lcc}
\hline
\textbf{Model} & \textbf{Config-Size} & \textbf{Accuracy (\%)} \\
\hline
GPT4o-mini & PubMed-MKG-8B & \textbf{73.92} \\
           & No-MKG \& CoT-8B & 66.69 \\
\hline
LLaMA 3.1 & PubMed-MKG-8B & 66.5 \\
          & No-MKG-8B & 62.6 \\
\hline
Mixtral & PubMed-MKG-8$\times$7B & 61.4 \\
        & No-MKG-8$\times$7B & 53.2 \\
\hline
GPT 3.5 & PubMed-MKG & 65.2 \\
        & No-MKG & 58.4 \\
\hline
\end{tabular}
\caption{\gls{myrag} performance across different LLM backbones on the MEDQA benchmark.}
\label{tab:medqa_llm_results}
\end{table}

\paragraph{Improving QA in Rapidly Changing Medical Domains with AMG-RAG.}

\begin{figure}[t]
    \centering
    \includegraphics[width=1\columnwidth]{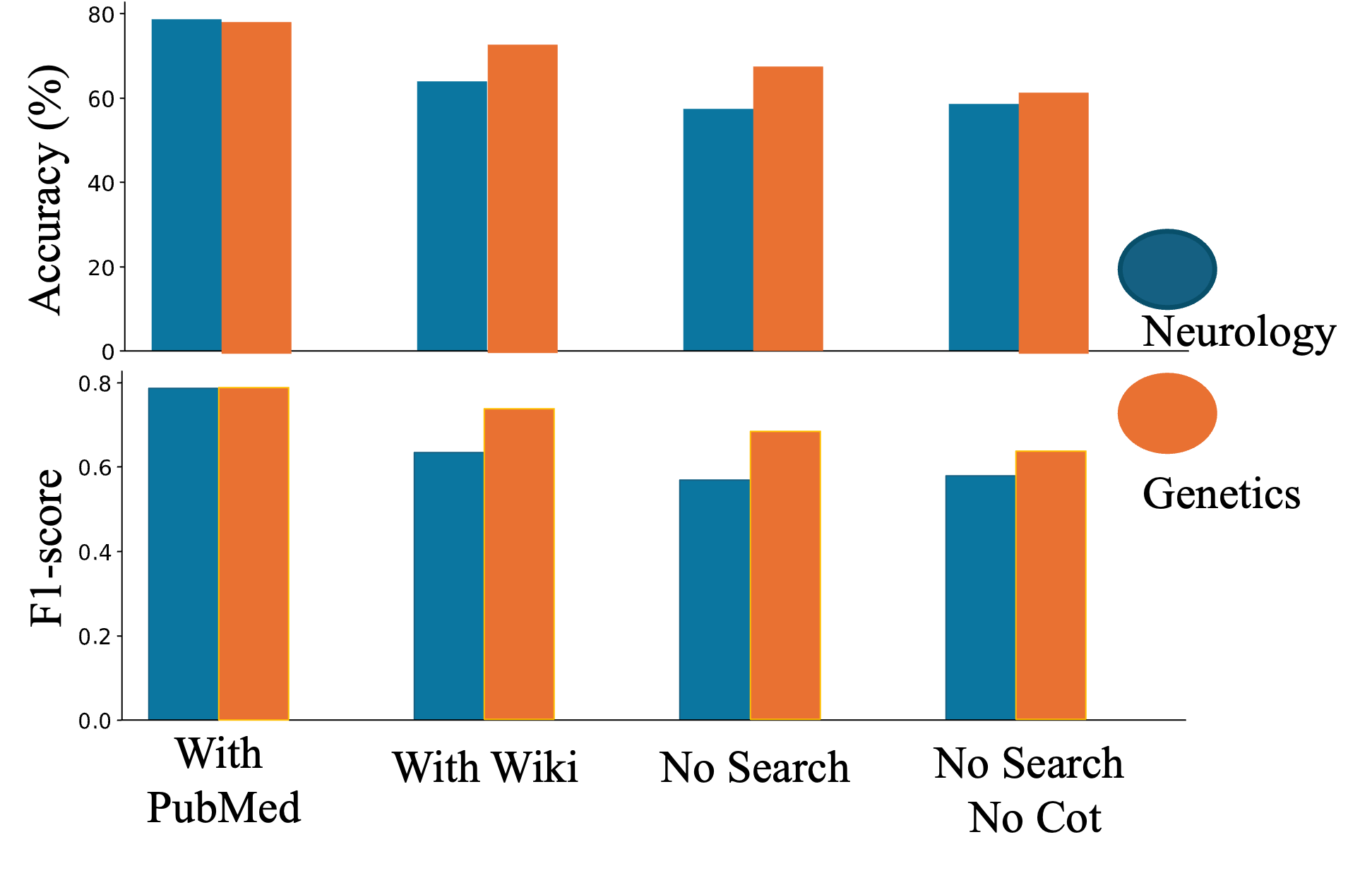}
    \caption{Performance comparison across different question domains in the Neurology and Genetics fields.}
    \label{fig:performance_clusters}
\end{figure}

Figure~\ref{fig:performance_clusters} shows \gls{myrag}'s superior performance in rapidly evolving subfields like {Neurology} and {Genetics}. This advantage stems from real-time PubMed integration during inference, combined with structured reasoning and knowledge graph grounding, enabling precise answers to complex medical questions with enhanced interpretability and trustworthiness.

\section{Conclusion}
We introduce \gls{myrag}, an advanced \gls{qa} system that dynamically constructs \gls{mkg} while integrating sophisticated structured reasoning for medical \gls{qa}. The system demonstrates significant improvements in accuracy and reasoning capabilities, particularly for medical question-answering tasks, outperforming other approaches of similar model size or 10 to 100 times larger. 
\section{Limitations}
Despite \gls{myrag}'s advancements, our approach has certain limitations. Firstly, it relies on external search tools which introduce latency during the creation of the \gls{mkg}. However, this occurs only once, when the \gls{mkg} is built from scratch for the first time. Additionally, while the system performs exceptionally well in medical domains, its applicability to non-medical tasks remains unexplored.

Another limitation is the need for structured, authoritative sources of medical knowledge. Currently, \gls{myrag} retrieves information from diverse sources, including research articles and medical textbooks. However, as emphasized in clinical decision-making, treatment guidelines serve as essential references for standardized diagnosis and treatment protocols \cite{hager2024evaluation}. Future work on \gls{myrag} should focus on integrating structured access to these sources to ensure compliance with evidence-based medicine.

\section{Ethics Statement}
The development of \gls{llm}s for medical \gls{qa} requires careful ethical consideration due to risks of inaccuracy and bias. Ensuring the reliability of retrieved content is crucial, especially when integrating external knowledge sources. To mitigate these risks, we implement a confidence scoring mechanism into the \gls{mkg} to validate the information. However, bias detection and mitigation remain active research areas.
\section*{Acknowledgements}

\bibliography{main}
\newpage
\appendix
\appendix
\section{Confidence Scoring for the Relationships in the MKG}
\label{app:confidence}
A confidence score, $s_{ij}$, is assigned to each inferred relationship, reflecting its strength and relevance. The scoring criteria are as follows:
\begin{itemize}
    \item \textbf{10}: The target is directly and strongly related to the item, with clear, unambiguous relevance.
    \item \textbf{7-9}: The target is moderately to highly relevant to the item but may have some ambiguity or indirect association.
    \item \textbf{4-6}: The target has some relevance to the item but is weak or only tangentially related.
    \item \textbf{1-3}: The target has minimal or no meaningful connection to the item.
\end{itemize}

\section{Evaluating the Accuracy and Robustness of the Medical Knowledge Graph}
\label{app:mkd-analysis}

The quality and reliability of the dynamically generated \gls{mkg} are critical for its effectiveness in enhancing medical \gls{qa} systems. To validate the accuracy, robustness, and usability of the \gls{mkg}, a structured evaluation involving expert \glspl{llm} in the medical domain, such as GPT medical model, was conducted. This section outlines the methodology used to evaluate the \gls{mkg}, emphasizing interpretability, clinical relevance, and robustness in real-world applications. Additionally, the role of medical experts in verifying the accuracy and applicability of the \gls{mkg} is discussed, underscoring the necessity of human expertise in validating AI-driven medical knowledge representations.

To assess accuracy and robustness, a two-phase evaluation process was employed. In the first phase, a group of expert \glspl{llm} specialists in medical domains reviewed a subset of the \gls{mkg}, including dynamically generated nodes, relationships, confidence scores, and summaries for various medical queries. They evaluated the accuracy of medical terms and concepts, the relevance of relationships between nodes, the reliability of node summaries, and the alignment of confidence scores with the perceived strength and reliability of the connections. Each \gls{llm} independently rated the graph components on a scale of 1 to 10. The results showed an average accuracy score of 8.9/10 for node identification, 8.8/10 for relationship relevance, and 8.5/10 for the clarity and precision of node summaries. Confidence scores generally aligned well with the \glspl{llm}'s assessments, as illustrated in Tables~\ref{tab:mkg_analysis_samples} and~\ref{tab:mkg_analysis_more_samples}, which highlight strong relationships across domains such as ophthalmology, cardiovascular treatments, and dermatology.

In the second phase, blind testing was conducted to evaluate usability and human-readability. Expert \glspl{llm} were tasked with answering complex medical queries requiring multi-hop reasoning, such as managing comorbidities or determining multi-drug treatment protocols. As shown in Table~\ref{tab:mkg_analysis_samples}, relationships such as the co-usage of Ketotifen and Fluorometholone for allergic conjunctivitis or Labetalol and Nitroglycerin for acute hypertension demonstrated the \gls{mkg}'s ability to model clinically relevant associations effectively. The \glspl{llm} achieved a 89\% accuracy rate in these test scenarios. Additionally, the \glspl{llm} rated the \gls{mkg} 9.4/10 for interpretability and usability, underscoring its strength in visually and contextually representing complex medical relationships.

To further ensure the clinical relevance and practical applicability of the \gls{mkg}, medical experts, including practicing physicians and clinical researchers, were involved in evaluating the generated relationships and summaries. Unlike \glspl{llm}, medical experts provided qualitative assessments, identifying potential discrepancies, overlooked nuances, and contextual dependencies that automated models might miss. The medical experts particularly assessed:
\begin{enumerate}
    \item The correctness and completeness of medical relationships, ensuring they align with established clinical knowledge and best practices.
    \item The validity of multi-hop reasoning paths, verifying whether inferred relationships reflected logical clinical decision-making processes.
    \item The utility of the \gls{mkg} in real-world medical applications, particularly in aiding diagnostic and treatment decision-making.
\end{enumerate}

The feedback from medical experts was instrumental in refining the graph, addressing inconsistencies, and enhancing the confidence scores to better reflect real-world medical reliability. Notably, medical expert ratings aligned well with \gls{llm} evaluations but provided deeper insights into the contextual limitations of the graph. For example, while \glspl{llm} accurately linked Diltiazem and Nitroglycerin in cardiovascular treatment, medical experts highlighted additional considerations such as contraindications in specific patient populations, which were subsequently incorporated into the \gls{mkg}.

The detailed evaluations in Tables~\ref{tab:mkg_analysis_samples} and~\ref{tab:mkg_analysis_more_samples} provide further insights into the graph’s performance across diverse medical domains. For instance, the accurate representation of relationships between beta-blockers like Labetalol and Propranolol or the integration of treatments such as Diltiazem and Nitroglycerin for cardiovascular care highlight the \gls{mkg}'s capacity to support intricate clinical decision-making.

These results confirm that the \gls{mkg} is both human-readable and usable by advanced \glspl{llm}, making it an invaluable tool for medical \gls{qa} and decision-making. The graph’s structured format, enriched with confidence scores and summaries, ensures a clear and interpretable representation of medical knowledge while enhancing the efficiency and accuracy of \gls{qa} systems in addressing real-world medical scenarios. Moreover, the involvement of medical experts in the evaluation process enhances the credibility of the \gls{mkg}, ensuring that AI-driven insights align with clinical expertise and practical healthcare applications.

\begin{table*}[ht]
\centering
\small

\resizebox{\textwidth}{!}{\begin{tabular}{|p{2cm}|p{3cm}|p{2cm}|p{3cm}|p{3cm}|p{3cm}|}
\hline
\textbf{Source Node} & \textbf{Relationship Type} & \textbf{Target Node} & \textbf{LLM Expert Analysis} & \textbf{Blind Analysis} & \textbf{Medical Expert Analysis}\\ \hline
Botulism & Directly related as it is the target concept. & Myasthenia gravis & Rated 9.2/10 for relevance and clinical importance, considered highly accurate. & Demonstrated effective multi-hop reasoning with a 92\% accuracy in identifying related conditions. & Rated 9.5/10 for relevance and accuracy, considered highly accurate.\\ \hline
Levodopa & Levodopa is a primary treatment for Parkinson's disease. & Parkinson's disease & Evaluated as highly reliable (9.6/10) for summarizing medical treatments and relationships. & Increases accuracy by 24\% in answering queries about Parkinson's treatments and comorbidities. & Rated 10/10 for relevance and accuracy, considered highly accurate.\\ \hline
Zidovudine & Zidovudine is an antiviral drug used for HIV treatment. & HIV/AIDS & Experts rated it 9.4/10 for interpretability, highlighting the clear representation of the relationship. & Provided contextually accurate responses regarding drug interactions and side effects in queries. & Rated 10/10 for relevance and accuracy, considered highly accurate.\\ \hline
Inhibition of thymidine synthesis & Cross-linking of DNA is directly related to thymidine synthesis as both involve nucleic acid metabolism. & Cross-linking of DNA & Rated 9.2/10 for relevance to nucleic acid metabolism and DNA replication. & Demonstrated high accuracy in answering multi-hop queries related to DNA synthesis pathways. & Rated 9/10 for relevance and accuracy, considered accurate.\\ \hline
Hyperstabilization of microtubules & Cross-linking of DNA can be related to the stabilization of microtubules. & Cross-linking of DNA & Rated 9.0/10 for highlighting structural modifications affecting cellular functions. & Increases the accuracy by 20\% in scenarios involving cellular structure interactions. & Rated 8/10 for moderated relevance.\\ \hline
Generation of free radicals & Free radicals can lead to oxidative damage, affecting DNA integrity and function. & Cross-linking of DNA & Rated 8.5/10 for its relevance to oxidative stress and DNA damage mechanisms. & Accurate in providing causal explanations for oxidative stress and DNA cross-linking. & Rated 7.5/10 for relevance.\\ \hline
Renal papillary necrosis & Allergic interstitial nephritis can lead to renal damage. & Allergic interstitial nephritis & Rated 9.0/10 for explaining the clinical progression of renal complications. & Effective in multi-hop reasoning for renal damage-related queries, achieving 91\% accuracy. & Rated 10/10 for relevance and accuracy, considered highly accurate.\\ \hline
\end{tabular}}
\caption{Examples from the Medical Knowledge Graph (MKG) with Expert and Blind Analysis (Part 1)}
\label{tab:mkg_analysis_samples}
\end{table*}

\begin{table*}[ht]
\centering
\small

\resizebox{\textwidth}{!}{\begin{tabular}{|p{2cm}|p{3cm}|p{2cm}|p{3cm}|p{3cm}|p{3cm}|}
\hline
\textbf{Source Node} & \textbf{Relationship Type} & \textbf{Target Node} & \textbf{LLM Expert Analysis} & \textbf{Blind Analysis} &\textbf{Medical Expert Analysis}\\ \hline
Ketotifen eye drops & Ketotifen eye drops are antihistamines used for allergic conjunctivitis, which may be used alongside Fluorometholone for managing eye allergies. & Fluorometholone eye drops & Rated 9.2/10 for relevance in managing allergic conjunctivitis. & Demonstrated 93\% accuracy in multi-hop reasoning for ophthalmological conditions. &Rated 10/10 for relevance and accuracy, considered highly accurate. \\ \hline
Ketotifen eye drops & Latanoprost eye drops are used to lower intraocular pressure in glaucoma, while Ketotifen treats allergic conjunctivitis. & Latanoprost eye drops & Rated 9.0/10 for distinct yet complementary roles in ophthalmology. & Effective in identifying separate ophthalmic applications with 92\% accuracy. & Rated 10/10 for relevance and accuracy, considered highly accurate.\\ \hline
Diltiazem & Nitroglycerin is relevant in discussions of cardiovascular treatments alongside Diltiazem. & Nitroglycerin & Rated 8.8/10 for contextual relevance to cardiovascular management. & Increases the accuracy for treatment-based queries by 20\%. & Rated 9.5/10 for relevance and accuracy, considered highly accurate.\\ \hline
Labetalol & Labetalol is closely related to Propranolol, both managing hypertension. & Propranolol & Rated 9.5/10 for direct relevance in cardiovascular treatment protocols. & Highly interpretable responses for hypertension management, with 95\% accuracy. &Rated 10/10 for relevance and accuracy, considered highly accurate. \\ \hline
Nitroglycerin & Nitroglycerin and Labetalol are often used in conjunction for managing hypertension and heart conditions. & Labetalol & Rated 8.7/10 for strong relevance in acute hypertension protocols. & Supported effective multi-drug therapy reasoning with 90\% accuracy. &Rated 9/10 for relevance and accuracy, considered highly accurate. \\ \hline
Nitroglycerin & Nitroglycerin is often used with Propranolol in managing cardiovascular conditions like hypertension and angina. & Propranolol & Rated 9.0/10 for its importance in cardiovascular multi-drug therapy. & Demonstrated robust performance in connecting treatment protocols, with 93\% query accuracy. &Rated 10/10 for relevance and accuracy, considered highly accurate. \\ \hline
Fluorometholone eye drops & Fluorometholone eye drops are corticosteroids that treat inflammation, complementing Ketotifen for allergic conjunctivitis. & Ketotifen eye drops & Rated 8.8/10 for their combined application in managing inflammation and allergies. & Improved query relevance for multi-drug therapy in eye care by 19\%. &Rated 9.5/10 for relevance and accuracy, considered highly accurate.\\ \hline
Lanolin & Lanolin is used for skin care, particularly for sore nipples during breastfeeding. & Fluorometholone eye drops & Rated 8.5/10 for highlighting non-overlapping yet clinically useful contexts. & Demonstrated effective differentiation of clinical uses with high interpretability. &Rated 9/10 for relevance and accuracy, considered highly accurate.\\ \hline
\end{tabular}}
\caption{Examples from the Medical Knowledge Graph (MKG) with Expert and Blind Analysis (Part 2)}
\label{tab:mkg_analysis_more_samples}
\end{table*}

\section{QA Samples with reasoning from MEDQA benchmark}
This section presents a set of \gls{qa} samples demonstrating the reasoning paths generated by our proposed \gls{myrag} model when applied to the MEDQA dataset. These examples highlight how the model retrieves relevant content, structures key information, and formulates reasoning to guide answer selection.

Table~\ref{table:search_guidance_1} provides an example of how the model processes a clinical case question related to the management of acute coronary syndrome (ACS). The search items retrieved for possible answer choices (e.g., Nifedipine, Enoxaparin, Clopidogrel, Spironolactone, Propranolol) are accompanied by key content excerpts relevant to their roles in ACS treatment. Additionally, the reasoning pathways illustrate how the model synthesizes evidence-based knowledge to justify the selection of the correct answer (Clopidogrel), while also explaining why the alternative options are not suitable. Additional examples are also provided in Tables ~\ref{table:search_guidance_2}, ~\ref{table:search_guidance_3}, and ~\ref{table:search_guidance_4}
\label{app:examples-medqa}
\begin{table*}[h!]
\centering
\begin{tabular}{|p{3cm}|p{4cm}|p{4cm}|}
\hline
\textbf{Search Item/ Question Options} & \textbf{Key Content Highlighted} & \textbf{Reasoning Guiding the Answer} \\ \hline
\textbf{Nifedipine} &
\textcolor{red}{Not typically used for acute coronary syndrome (ACS). Associated with reflex tachycardia.} &
Nifedipine is a calcium channel blocker effective for hypertension but does not address the antiplatelet needs of ACS patients. \\ \hline
\textbf{Enoxaparin} &
\textcolor{blue}{Used for anticoagulation in ACS but mainly during hospitalization.} &
Enoxaparin is not continued after discharge when aspirin and another antiplatelet drug are prescribed. \\ \hline
\textbf{Clopidogrel} &
\textcolor{green}{Standard for dual antiplatelet therapy (DAPT) in ACS, especially post-percutaneous coronary intervention (PCI).} &
Clopidogrel complements aspirin in preventing thrombotic events post-angioplasty. Its use is supported by evidence-based guidelines. \\ \hline
\textbf{Spironolactone} &
\textcolor{orange}{Useful in heart failure or reduced ejection fraction but not indicated for ACS management when EF is normal.} &
This patient's EF is 58\%, so spironolactone is not necessary. Focus should be on antiplatelet therapy. \\ \hline
\textbf{Propranolol} &
\textcolor{purple}{Effective for reducing myocardial oxygen demand but not part of standard DAPT.} &
While beneficial for stress-related heart issues, it does not address thrombotic risks in ACS management. \\ \hline
\end{tabular}
\caption{Examples of Summary of search items for the question "A 65-year-old man is brought to the emergency department 30 minutes after the onset of acute chest pain. He has hypertension and asthma. Current medications include atorvastatin, lisinopril, and an albuterol inhaler. He appears pale and diaphoretic. His pulse is 114/min, and blood pressure is 130/88 mm Hg. An ECG shows ST-segment depressions in leads II, III, and aVF. Laboratory studies show an increased serum troponin T concentration. The patient is treated for acute coronary syndrome and undergoes percutaneous transluminal coronary angioplasty. At the time of discharge, echocardiography shows a left ventricular ejection fraction of 58\%. In addition to aspirin, which of the following drugs should be added to this patient's medication regimen?" and Their Influence on the Correct Answer (Clopidogrel) and the reasoning paths}
\label{table:search_guidance_1}
\end{table*}

\begin{table*}[h!]
\centering
\begin{tabular}{|p{3cm}|p{4cm}|p{4cm}|}
\hline
\textbf{Search Item/ Question Options} & \textbf{Key Content Highlighted} & \textbf{Reasoning Guiding the Answer} \\ \hline

\textbf{A history of stroke or venous thromboembolism} &
\textcolor{red}{Contraindicated for hormonal contraceptives due to increased risk of thrombosis.} &
Copper IUDs do not carry the same thrombotic risk, making this option irrelevant for contraindication in IUD placement. \\ \hline

\textbf{Current tobacco use} &
\textcolor{blue}{Increases cardiovascular risk with hormonal contraceptives but not with copper IUDs.} &
Tobacco use does not contraindicate IUD placement, though it may influence other contraceptive choices. \\ \hline

\textbf{Active or recurrent pelvic inflammatory disease (PID)} &
\textcolor{green}{Direct contraindication for IUD placement due to the risk of exacerbating infection and complications.} &
Insertion of an IUD can worsen active PID, leading to infertility or other severe complications. \\ \hline

\textbf{Past medical history of breast cancer} &
\textcolor{orange}{Contraindicates hormonal contraceptives, but copper IUDs are considered safe.} &
This option does not contraindicate copper IUD placement, as it is non-hormonal and unrelated to breast cancer. \\ \hline

\textbf{Known liver neoplasm} &
\textcolor{purple}{Contraindicates hormonal contraceptives but not copper IUDs.} &
Copper IUDs are safe for patients with liver neoplasms as they are free of systemic hormones. \\ \hline

\end{tabular}
\caption{Examples of Summary of Search Items for the Question "A 37-year-old-woman presents to her primary care physician requesting a new form of birth control. She has been utilizing oral contraceptive pills (OCPs) for the past 8 years, but asks to switch to an intrauterine device (IUD). Her vital signs are: blood pressure 118/78 mm Hg, pulse 73/min and respiratory rate 16/min. She is afebrile. Physical examination is within normal limits. Which of the following past medical history statements would make copper IUD placement contraindicated in this patient?" and Their Influence on the Correct Answer (Active or recurrent pelvic inflammatory disease (PID)) and the Reasoning Paths}
\label{table:search_guidance_2}
\end{table*}

\begin{table*}[h!]
\centering
\begin{tabular}{|p{3cm}|p{4cm}|p{4cm}|}
\hline
\textbf{Search Item/ Question Options} & \textbf{Key Content Highlighted} & \textbf{Reasoning Guiding the Answer} \\ \hline

\textbf{Dementia} &
\textcolor{red}{Typically presents as a gradual decline in cognitive function.} &
The sudden onset of symptoms after surgery and acute confusion makes dementia less likely. \\ \hline

\textbf{Alcohol withdrawal} &
\textcolor{blue}{Requires significant and sustained alcohol use to cause withdrawal symptoms.} &
The patient's weekly consumption of one to two glasses of wine is insufficient to support this diagnosis. \\ \hline

\textbf{Opioid intoxication} &
\textcolor{purple}{Oxycodone can cause sedation and confusion, but stable vital signs and lack of severe respiratory depression are inconsistent.} &
While oxycodone use is relevant, the observed fluctuating agitation and impulsivity are more consistent with delirium. \\ \hline

\textbf{Delirium} &
\textcolor{green}{Characterized by acute changes in attention and cognition with fluctuating levels of consciousness.} &
The patient's recent surgery, medication use, and fluctuating symptoms align strongly with a diagnosis of delirium. \\ \hline

\textbf{Urinary tract infection (UTI)} &
\textcolor{orange}{Confusion in elderly patients can result from UTIs, but a normal urine dipstick test does not support this.} &
The absence of urinary findings on examination makes UTI less likely as the cause of symptoms. \\ \hline

\end{tabular}
\caption{Examples of Search Items for the Question: "Six days after undergoing surgical repair of a hip fracture, a 79-year-old woman presents with agitation and confusion. Which of the following is the most likely cause of her current condition?" and Their Influence on the Correct Answer (Delirium) and the Reasoning Paths.}
\label{table:search_guidance_3}
\end{table*}

\begin{table*}[h!]
\centering
\begin{tabular}{|p{3cm}|p{4cm}|p{4cm}|}
\hline
\textbf{Search Item/ Question Options} & \textbf{Key Content Highlighted} & \textbf{Reasoning Guiding the Answer} \\ \hline

\textbf{Primary spermatocyte} &
\textcolor{red}{Nondisjunction events during meiosis I often occur at this stage, leading to chromosomal abnormalities.} &
Klinefelter syndrome (47,XXY) is typically due to nondisjunction during meiosis, specifically at this stage. \\ \hline

\textbf{Secondary spermatocyte} &
\textcolor{blue}{Meiosis II occurs here, dividing chromosomes into haploid cells, but errors at this stage are less likely to lead to 47,XXY.} &
The chromosomal abnormality associated with Klinefelter syndrome usually arises before this stage. \\ \hline

\textbf{Spermatid} &
\textcolor{purple}{Spermatids are post-meiotic cells where genetic material is already finalized.} &
Errors at this stage would not result in a cytogenetic abnormality like 47,XXY. \\ \hline

\textbf{Spermatogonium} &
\textcolor{green}{Errors here affect the germline but are less likely to cause specific meiotic nondisjunction errors.} &
While germline mutations can occur, meiotic nondisjunction leading to Klinefelter syndrome occurs later. \\ \hline

\textbf{Spermatozoon} &
\textcolor{orange}{These are fully mature sperm cells that inherit abnormalities from earlier stages.} &
By this stage, chromosomal errors have already been established. \\ \hline

\end{tabular}
\caption{Examples of Search Items for the Question: "A 29-year-old man with infertility, tall stature, gynecomastia, small testes, and an elevated estradiol:testosterone ratio is evaluated. Genetic studies reveal a cytogenetic abnormality inherited from the father. At which stage of spermatogenesis did this error most likely occur?" and Their Influence on the Correct Answer (Primary spermatocyte) and the Reasoning Paths.}
\label{table:search_guidance_4}
\end{table*}

\section{Implementation Details for Dataset Ingestion and Vector Database}
\label{app:vector_db}

This section outlines the pipeline for dataset ingestion and vector database creation for efficient medical question-answering. The process involves document chunking, embedding generation, and storage in a vector database to facilitate semantic retrieval.

\subsection{Dataset Processing and Chunking}
The dataset, sourced from medical textbooks in the MEDQA benchmark, is provided in plain text format. Each document is segmented into smaller chunks with a maximum size of 512 tokens and a 100-token overlap. This overlap ensures context preservation across chunk boundaries, supporting multi-hop reasoning for long documents.

\subsection{Embedding Model and Vector Storage}
The system utilizes the \textbf{SentenceTransformer} model, specifically \texttt{all-mpnet-base-v2}, for generating dense vector representations of text chunks and queries. To optimize storage and retrieval, the embeddings are indexed in the \textbf{Chroma} vector database. Metadata, such as document filenames and chunk IDs, is also stored to maintain document traceability.

\subsection{Batch Processing and Vector Database Population}
To manage memory efficiently during ingestion, document chunks are processed in batches of up to 10,000. This ensures a smooth ingestion pipeline while preventing memory overflow. Each processed file is logged to avoid redundant computations, and error handling mechanisms are in place to manage failed processing attempts.

\subsection{Query Answering Workflow}
For retrieval, user queries (e.g., \textit{"What are the symptoms of drug-induced diabetes?"}) are embedded using the \texttt{all-mpnet-base-v2} model. The top-ranked relevant chunks are retrieved based on their semantic similarity to the query using Chroma's similarity search mechanism. The system retrieves the top \(k\) relevant passages, which can be further processed in downstream QA models.

\subsection{Key Configuration Details}
The system is configured with the following parameters:
\begin{itemize}
    \item \textbf{Embedding Model:} \texttt{all-mpnet-base-v2} from SentenceTransformer.
    \item \textbf{Vector Database:} Chroma, stored persistently on disk for reusability.
    \item \textbf{Chunk Size:} 512 tokens per chunk, with a 100-token overlap for contextual consistency.
    \item \textbf{Batch Size:} Up to 10,000 chunks per batch to optimize ingestion efficiency.
\end{itemize}

\subsection{Implementation and System Execution}
The ingestion and query process is implemented using Python, leveraging \texttt{sentence-transformers} for embeddings and \texttt{Chroma} for vector storage. The ingestion pipeline reads and processes text files, splits them into chunks, generates embeddings, and stores them efficiently in the vector database. The querying process retrieves the top \(k\) most relevant text chunks to respond to user queries.

\section{Components Definition }
\subsection{Neo4j}

As data complexity increases, traditional relational databases struggle with highly interconnected datasets where relationships are crucial. Graph databases, like Neo4j, address this challenge by efficiently modeling and processing complex, evolving data structures using nodes, relationships, and properties \cite{besta2023demystifying}. 

Neo4j, an open-source NoSQL graph database, enables constant-time traversals by explicitly storing relationships, making it ideal for large-scale applications such as social networks, recommendation systems, and biomedical research. Unlike relational models, Neo4j avoids costly table joins and optimizes deep relationship queries, enhancing scalability and performance \cite{besta2023demystifying}. 

Neo4j's architecture is centered around the property graph model, which includes\cite{huang2013research}:

\begin{itemize}
    \item \textbf{Nodes}: Entities representing data points.
    \item \textbf{Relationships}: Directed, named connections between nodes that define how entities are related.
    \item \textbf{Properties}: Key-value pairs associated with both nodes and relationships, providing additional metadata.
\end{itemize}

This model allows for intuitive representation of complex data structures and supports efficient querying and analysis. The system's internal mechanisms facilitate rapid traversal of relationships, enabling swift query responses even in large datasets \cite{huang2013research}.

\paragraph{Does Neo4j-Based Storage scale well?} Neo4j's scalability for our medical knowledge graph storage is strategically robust, offering several key advantages for large-scale, relationship-intensive medical data. Its graph-based architecture is particularly well-suited for handling highly interconnected medical knowledge networks, supporting horizontal scaling that enables efficient performance even as the knowledge base grows. The cloud-based accessibility further enhances the framework's flexibility, allowing seamless knowledge sharing and distributed access without local storage constraints.

\paragraph{How large were the knowledge graphs?}
Our automatically constructed medical knowledge graphs demonstrate significant complexity and depth, comprising approximately 76,681 nodes and 354,299 edges. These nodes encompass a comprehensive range of medical entities including diseases, symptoms, treatments, drugs, anatomical structures, and clinical findings, all interconnected through semantically meaningful, typed relationships. This substantial scale not only reflects the intricate nature of medical knowledge but also enables more nuanced, multi-hop reasoning capabilities across diverse medical queries. The graph's architecture allows for dynamic expansion and refinement, ensuring that the knowledge representation remains both comprehensive and adaptable to emerging medical research and understanding.

\section{Additional Results}
\begin{table*}[t]
\centering
\small
\begin{tabular}{p{0.1\linewidth} p{0.85\linewidth}}
\toprule
\textbf{Example 1} & 
\textbf{Question:} A 29-year-old man presents with infertility. He has been trying to conceive for over 2 years. His wife has no fertility issues. Exam shows tall stature, long limbs, sparse body hair, gynecomastia, and small testes. Labs reveal elevated FSH and a high estradiol:testosterone ratio. Cytogenetic analysis indicates a chromosomal abnormality. If inherited from the father, during which stage of spermatogenesis did this error most likely occur? \\
& \textbf{Choices:} A: Primary spermatocyte, B: Secondary spermatocyte, C: Spermatid, D: Spermatogonium, E: Spermatozoon \\
& \textbf{Answer:} A (Primary spermatocyte) \\
& \textbf{Reasoning:} This corresponds to an error in meiosis I during the father's spermatogenesis, consistent with Klinefelter syndrome due to paternal nondisjunction. \\
& \textbf{Retrieved Papers:} 
1) Black et al., *The Genetic Landscape of Male Factor Infertility*, Uro, 2025. 
2) Niyaz et al., *Chromosome Disorders in Sperm Anomalies*, 2025. 
3) Leslie et al., *MNS1 variant and Male Infertility*, EJHG, 2020. \\
\midrule
\textbf{Example 2} & 
\textbf{Question:} A 23-year-old woman is referred for genetic counseling after her brother is diagnosed with hereditary hemochromatosis. She is asymptomatic and her labs are normal. Which gene mutation is most consistent with hereditary hemochromatosis? \\
& \textbf{Choices:} A: BCR-ABL, B: BRCA, C: FA, D: HFE, E: WAS \\
& \textbf{Answer:} D (HFE gene) \\
& \textbf{Reasoning:} Most hereditary hemochromatosis cases in Northern European populations are caused by HFE mutations (C282Y, H63D). Even asymptomatic individuals with normal iron studies should be screened if they have an affected first-degree relative. \\
& \textbf{Retrieved Papers:}
1) Delatycki \& Allen, *Population Screening for HH*, Genes, 2024. 
2) Lou et al., *Utility of Iron Indices in HH Genotyping*, Clin. Biochem., 2025. 
3) Lucas et al., *HFE Genotypes and Outcomes*, BMJ Open, 2024. \\
\bottomrule
\end{tabular}
\caption{Examples of AMG-RAG-generated answers with structured reasoning and citation-based grounding for clinical QA.}
\label{tab:amg_rag_examples}
\end{table*}

\end{document}